\documentclass[10pt,twocolumn,letterpaper]{article}

\usepackage{cvpr}              % To produce the CAMERA-READY version
\definecolor{cvprblue}{rgb}{0.21,0.49,0.74}
\usepackage[pagebackref,breaklinks,colorlinks,allcolors=cvprblue]{hyperref}

\def\paperID{} % *** Enter the Paper ID here
\def\confName{CVPR}
\def\confYear{2025}

\title{ Evaluating Cell Type Inference in Vision Language Models Under Varying Visual Context}

\author{Samarth Singhal\\
University of North Dakota\\
{\tt\small samarth.singhal@und.edu}
\and
Sandeep Singhal\\
University of North Dakota\\
{\tt\small sandeep.singhal@und.edu}
}

\begin{document}
\maketitle
\begin{abstract}
Vision-Language Models (VLMs) have rapidly advanced alongside Large Language Models (LLMs). This study evaluates the capabilities of prominent generative VLMs, such as GPT-4.1 and Gemini 2.5 Pro, accessed via APIs, for histopathology image classification tasks, including cell typing. Using diverse datasets from public and private sources, we apply zero-shot and one-shot prompting methods to assess VLM performance, comparing them against custom-trained Convolutional Neural Networks (CNNs). Our findings demonstrate that while one-shot prompting significantly improves VLM performance over zero-shot ($p \approx 1.005 \times 10^{-5}$ based on Kappa scores), these general-purpose VLMs currently underperform supervised CNNs on most tasks. This work underscores both the promise and limitations of applying current VLMs to specialized domains like pathology via in-context learning. All code and instructions for reproducing the study can be accessed from the repository \href{https://www.github.com/a12dongithub/VLMCCE}{GitHub}.  
\end{abstract}    
\section{Introduction}
\label{sec:intro}

Recent advancements in Vision-Language Models (VLMs) \cite{radford2021learning, https://doi.org/10.48550/arxiv.2204.14198}, which process both visual and textual information, have spurred significant progress. Early foundational VLMs like CLIP \cite{radford2021learning} focused primarily on learning powerful joint image-text representations. More recently, a new generation of sophisticated \textbf{VLMs}, including models like GPT-4V \cite{openai2023gpt4}, Gemini \cite{google2023gemini}, and InternVL \cite{chen2024internvl}, have demonstrated remarkable capabilities in understanding and generating content based on complex multimodal inputs. Crucially, these advanced VLMs can follow intricate instructions interleaved with images and text, enabling direct application to tasks like visual question answering and prompted classification.

Deep learning has revolutionized computational pathology, yet state‑of‑the‑art classifiers require thousands of labeled examples and costly re‑training for each new biomarker, stain, or organ~\cite{Madabhushi2016}. Despite these gains, three persistent bottlenecks hamper widespread adoption.

\textbf{Data scarcity:} Every new diagnostic question demands a fresh cohort of meticulously annotated images \cite{Gessain2025}, an onerous burden for already over‑extended pathology services.

\textbf{Compute barriers:} Fine‑tuning large vision backbones for each task is compute‑intensive and often beyond the reach of smaller research groups or hospitals~\cite{Morales2021, Bessen2025}.

\textbf{Limited knowledge integration:} Classical CNNs operate solely on pixels, failing to exploit the rich textual and clinical context \cite{https://doi.org/10.48550/arxiv.1907.09478} that accompanies histology slides. 

\begin{figure}[t]

  \centering
   \includegraphics[width=0.8\linewidth]{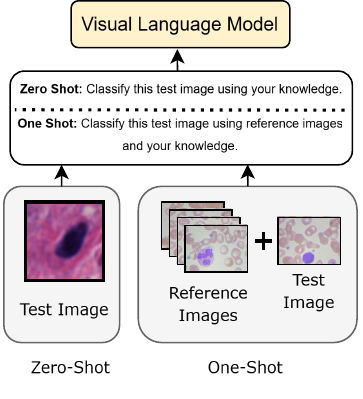}

   \caption{Evaluation Pipeline for Generative VLMs}
   \label{fig:onecol}
\end{figure}

Recent work in \emph{in‑context learning} (ICL) \cite{Ferber2024} shows that large language models can generalize from only a handful of examples provided at inference time~\cite{https://doi.org/10.48550/arxiv.2005.14165, https://doi.org/10.48550/arxiv.2311.16452}. Building on these advances, we adapt the ICL paradigm to pathology images, specifically for cell and tissue classification. We test two prompt regimes—\textbf{zero-shot} and \textbf{one-shot}—using state-of-the-art \textbf{Vision-Language Models (VLMs)} such as GPT-4.1, Gemini 2.5 Pro, InternVL 3, and GPT-4.1 Mini, selected for their ability to process interleaved image-text prompts and follow instructions. We also evaluate the medical domain-specific VLM, \emph{LLaVA-Med} \cite{li2023llavamed} in our study. Our experiments extend ICL to vision–language pathology tasks and provide the first head-to-head comparison of these current general-purpose \textbf{VLMs} under both settings. Crucially, we treat these models as \emph{unmodified} foundations—\emph{not} clinical tools—to (i) quantify their "as is" performance via prompting, and (ii) identify gaps that a dedicated medical \textbf{VLM} and/or Agent \cite{ https://doi.org/10.48550/arXiv.2304.03442} might need to address.

\section{Methods}
\subsection{Evaluation Pipeline}
Fig.~\ref{fig:onecol} schematically illustrates the ordering of images and text inside every inference prompt. Approximately 200 images were used per class for testing across all datasets. 
Each call embeds both modalities in a single message constructed in one of two regimes:

\medskip
\noindent\textbf{Zero-shot:} The prompt contains only the query image followed by  

\noindent\texttt{\small ``Classify this test image based on your knowledge alone. Choices: $\langle c_1\rangle,\dots,\langle c_K\rangle$. Reply with the class name in one or two words maximum.''}

\noindent This measures the innate pathology knowledge of an unmodified VLM.

\medskip
\noindent\textbf{One-shot:} For every class $c_k$ we first insert exactly one reference image and its caption  

\noindent\texttt{\small ``This image is class: $c_k$''}.  

\noindent After the $K$ exemplars we append the query image and instruct  

\noindent\texttt{\small ``Now classify this test image using reference images and your knowledge. Choices: $\langle c_1\rangle,\dots,\langle c_K\rangle$. Reply with the class name in one or two words maximum.''}

\medskip
The explicit enumeration of class names enforces a closed-set decision while enabling the model to combine visual cues from the exemplars with its pretrained medical priors.

\subsection{Datasets}
Our benchmarks draw on a diverse collection of histopathology image sets that vary in tissue origin, staining protocol, and annotation granularity. Table~\ref{tab:datasets} summarizes dataset specifics, including class counts and average image dimensions (in pixels).

% \textbf{BCCD} \cite{mooney2018bloodcells}\,—\,peripheral contains 12,500 augmented images of blood cells  blood–smear images, \textbf{Matek \etal} \cite{matek2021dataset}\,—\,bone-marrow cytology images of May-Grünwald-Giemsa/Pappenheim stain, were used \emph{as published}; each already provides class labels (four leukocyte classes and ten bone-marrow cell classes, respectively), so no additional preprocessing was required.  

We utilized several existing datasets, all used \emph{as published} without requiring additional preprocessing. These include: \textbf{Blood Cell Count and Detection  (BCCD)} \cite{mooney2018bloodcells}\,—\,providing 12,500 augmented peripheral blood smear images with four leukocyte class labels; \textbf{Matek \etal} \cite{matek2021dataset}\,—\,containing bone-marrow cytology images (May-Grünwald-Giemsa/Pappenheim stain) with ten bone-marrow cell class labels; and the \textbf{Meng \etal} tissue dataset \cite{h2qw97-18} (time-stretch microscopy), which ships pre-segmented images and three phenotype labels.

The \textbf{Private Bladder} dataset (collected under IRB approval) includes 3 Hematoxylin and Eosin (H\&E) whole-slide images (WSIs) at 40x magnification from 3 deidentified patients with urothelial carcinoma. Each WSI represents a unique patient and was assigned a distinct tumor aggressiveness grade (1, 2, or 3) by a certified pathologist, who delineated broad tumor and non-tumor regions. This dataset supports two derived tasks: (1) \textbf{Tumor Grading Classification:} using patches sampled from regions corresponding to the pathologist-assigned grade for each WSI. (2) \textbf{Cell-Type Classification:} distinguishing tumor vs. non-tumor nuclei. For this, nuclei within the pathologist-delineated regions were first segmented using QuPath \cite{Bankhead2017} (hematoxylin channel analysis), and then 10,000, 250x250 pixel patches centered on these nuclei were extracted \textit{per class} (i.e., 10k tumor nuclei patches, \~10k non-tumor nuclei patches). 

\textbf{PanNuke} \cite{gamper2019pannuke} contains H\&E patches spanning $19$ tissue types. We leveraged its nucleus segmentation masks to isolate individual cells: bounding boxes were placed around each mask and cropped to fixed-size patches, yielding balanced sets of neoplastic, inflammatory, connective, dead, and epithelial nuclei.

% Finally, we include the \textbf{LSMC} tissue dataset \cite{LSMC2024} (origin \textit{[tissue]}, stain \textit{[stain]}), which ships pre-segmented images and three phenotype labels; it, too, required no preprocessing.

Together these collections cover cytology, surgical pathology, and nuclear phenotyping under multiple class cardinalities, providing a rigorous test bed for zero-shot and one-shot vision–language inference.

\begin{table}[htbp]
\centering

\vspace{2mm}
\resizebox{\linewidth}{!}{%
\begin{tabular}{lccc}
\toprule
Dataset & Classes & Tissue & Avg. Pixel Size \\
\midrule
Bladder Tumor & 2  & Bladder & 64x64\\ 
Meng \etal \cite{h2qw97-18}& 3 & Blood and Breast &  128x128 \\
Bladder Grade & 3 & Bladder & 280x280\\ 
BCCD \cite{mooney2018bloodcells}& 4 & Blood & 320x240\\ 
PanNuke \cite{gamper2019pannuke}& 5 & Various Tissues & 46x46\\ 
Matek \etal \cite{matek2021dataset}& 10 & Bone Marrow & 250x250\\ 
\bottomrule
\end{tabular}}
\caption{Dataset Overview}
\label{tab:datasets}
\vspace{-2mm}
\end{table}

\subsection{Models}

We evaluated a spectrum of models ranging from a traditional Convolutional Neural Network (CNN) \cite{https://doi.org/10.48550/arxiv.1511.08458} baseline to several state-of-the-art Vision-Language Models (VLMs), applying the zero-shot and one-shot prompting strategies described in Tab.~\ref{tab:model_comparison_final_highlighted} without any model fine-tuning for the VLMs.

\medskip % Adds a little vertical space
\noindent\textbf{Custom CNN:} As a baseline, we trained a standard CNN with three sequential Conv3x3-ReLU-MaxPool blocks followed by a dense classifier head. This model was trained conventionally on the respective dataset tasks, using training sets containing, on average, at least 6000 images per class where applicable.

\medskip
\noindent\textbf{Vision-Language Models (VLMs):} We assessed the following large pre-trained models accessed via their APIs or standard implementations:
\begin{itemize}
    \item \textbf{Gemini 2.5 Pro \cite{google2023gemini}:} A leading closed-source multimodal model from Google, recognized for its number one spot on huggingface VLM leaderboard \cite{duan2024vlmevalkit} benchmarks at the time of evaluation.
    \item \textbf{GPT-4.1 \cite{openai2023gpt4}:} The latest flagship large multimodal model release from OpenAI.
    \item \textbf{GPT-4.1 Mini:} A smaller, potentially more efficient variant of OpenAI's GPT-4.1 model.
    \item \textbf{InternVL \cite{chen2024internvl}:} A prominent open-source VLM known for strong performance across various vision-language tasks.
    \item \textbf{LLaVA-Med \cite{li2023llavamed}:} A VLM specifically adapted for the general medical domain, evaluated to assess potential benefits of domain-specific pre-training. % Included as mentioned in intro
\end{itemize}
These VLMs were used directly as foundation models to evaluate their inherent capabilities for pathology image understanding through in-context learning prompts.

\subsection{Evaluation Metrics}
We quantified model performance using three established classification metrics: Accuracy, Macro-averaged F1-score, and Cohen's Kappa coefficient.

\textbf{Accuracy} measures the overall proportion of correctly classified instances:
\begin{equation}
\text{Accuracy} = \frac{\text{Number of Correct Predictions}}{\text{Total Number of Predictions}}
\end{equation}

\textbf{Macro-averaged F1-score} computes the harmonic mean of precision and recall independently per class, then averages these scores uniformly across all classes. This approach gives equal importance to each class irrespective of class distribution:
\begin{equation}
\text{F1}_{macro} = \frac{1}{N}\sum_{i=1}^{N}2 \cdot \frac{\text{Precision}_i \cdot \text{Recall}_i}{\text{Precision}_i + \text{Recall}_i}
\end{equation}
where $N$ represents the total number of classes.

\textbf{Cohen's Kappa} evaluates the classification agreement adjusted for chance:
\begin{equation}
\kappa = \frac{p_o - p_e}{1 - p_e}
\end{equation}
where $p_o$ is the observed accuracy and $p_e$ is the expected accuracy due purely to chance agreement.

Together, these metrics provide a comprehensive quantitative assessment of each model’s predictive reliability, class-balanced accuracy, and consistency beyond random chance, ensuring rigorous comparative analysis.

\section{Results}
\label{sec:Results}

\newcommand{\best}[1]{\textbf{#1}} % Command for category best
\newcommand{\overallbest}[1]{\textbf{\underline{#1}}} % Command for overall best
% The updated table code with highlighting:
\begin{table*}[t!]
\centering

\vspace{1mm}
\resizebox{\linewidth}{!}{% Adjust table width to fit text width
% l = Model column | c | c | c | ... = columns with vertical lines
\begin{tabular}{l|ccc|ccc|ccc|ccc|ccc|ccc} % Added vertical lines
\toprule
 % Header Row 1: Dataset Name & Class Count (Spanning 3 metric columns each)
 & \multicolumn{3}{c|}{Bladder Tumor (n=2)} % Added | for line
 & \multicolumn{3}{c|}{Bladder Grade (n=3)} % Added | for line
 & \multicolumn{3}{c|}{Meng \cite{h2qw97-18} \etal(n=3)} % Added | for line
 & \multicolumn{3}{c|}{BCCD \cite{mooney2018bloodcells} (n=4)} % Updated name, Added | for line
 & \multicolumn{3}{c|}{PanNuke \cite{gamper2019pannuke} (n=5)} % Added | for line
 & \multicolumn{3}{c}{Matek \cite{matek2021dataset} \etal(n=10)} \\ % Updated name, no final | needed
 % Header Row 2: Dimensions (Normal font size, spanning 3 metric columns each)
 & \multicolumn{3}{c|}{(64x64)} % Removed \scriptsize, Added |
 & \multicolumn{3}{c|}{(280x280)} % Removed \scriptsize, Added |
 & \multicolumn{3}{c|}{(128x128)} % Removed \scriptsize, Added |
 & \multicolumn{3}{c|}{(320x240)} % Removed \scriptsize, Added |
 & \multicolumn{3}{c|}{(46x46)} % Removed \scriptsize, Added |
 & \multicolumn{3}{c}{(250x250)} \\ % Removed \scriptsize
 % Header Row 3: Metrics (repeated), with rules grouping them under datasets
 \cmidrule(lr){2-4} \cmidrule(lr){5-7} \cmidrule(lr){8-10} \cmidrule(lr){11-13} \cmidrule(lr){14-16} \cmidrule(lr){17-19}
Model & Acc & F1 & $\kappa$ & Acc & F1 & $\kappa$ & Acc & F1 & $\kappa$ & Acc & F1 & $\kappa$ & Acc & F1 & $\kappa$ & Acc & F1 & $\kappa$ \\
\midrule
\multicolumn{19}{c}{\textbf{Supervised}} \\
% CNN row - Columns reordered: Bladder Tumor | Bladder Grade | Meng | BCCD | PanNuke | Matek
% CNN is always best in its category. Underline if also overall best.
CNN & \overallbest{0.960} & \overallbest{0.960} & \overallbest{0.920} & \overallbest{0.990} & \overallbest{0.990} & \overallbest{0.985} & \overallbest{0.887} & \overallbest{0.886} & \overallbest{0.830} & \best{0.711} & \best{0.710} & \best{0.615} & \overallbest{0.552} & \overallbest{0.518} & \overallbest{0.440} & \overallbest{0.779} & \overallbest{0.734} & \overallbest{0.748} \\
% ResNet row removed
\midrule
\multicolumn{19}{c}{\textbf{Zero Shot}} \\
% Apply \best{} to the max in each ZS column group.
GPT 4.1 & 0.502 & 0.343 & 0.005 & \best{0.360} & \best{0.239} & \best{0.040} & 0.347 & 0.203 & 0.020 & \best{0.766} & \best{0.755} & \best{0.688} & \best{0.271} & 0.268 & \best{0.089} & 0.353 & 0.327 & 0.280 \\
GPT 4.1 Mini & 0.527 & 0.406 & 0.055 & 0.257 & 0.206 & -0.090 & 0.328 & \best{0.328} & -0.007 & 0.646 & 0.617 & 0.528 & 0.203 & 0.207 & 0.004 & 0.234 & 0.197 & 0.160 \\
Gemini 2.5 Pro & 0.420 & 0.333 & -0.004 & 0.170 & 0.167 & -0.004 & 0.088 & 0.124 & -0.004 & 0.723 & 0.693 & 0.635 & 0.162 & 0.194 & 0.066 & \best{0.386} & \best{0.410} & \best{0.331} \\
InternVL3 & \best{0.618} & \best{0.594} & \best{0.235} & 0.337 & 0.175 & 0.005 & \best{0.378} & 0.286 & \best{0.068} & 0.636 & 0.639 & 0.515 & 0.268 & \best{0.268} & 0.085 & 0.172 & 0.158 & 0.095 \\
\midrule
\multicolumn{19}{c}{\textbf{One Shot}} \\
% Apply \best{} or \overallbest{} to the max in each OS column group.
GPT 4.1 & 0.637 & 0.636 & 0.275 & 0.595 & 0.550 & 0.393 & 0.323 & 0.287 & 0.004 & \overallbest{0.920} & \overallbest{0.920} & \overallbest{0.893} & 0.435 & 0.429 & 0.294 & \best{0.489} & \best{0.471} & \best{0.422} \\
GPT 4.1 Mini & 0.527 & 0.497 & 0.055 & \best{0.677} & \best{0.666} & \best{0.515} & 0.352 & 0.287 & 0.028 & 0.797 & 0.784 & 0.730 & \best{0.499} & \best{0.504} & \best{0.374} & 0.411 & 0.400 & 0.332 \\
Gemini 2.5 Pro & 0.502 & 0.499 & 0.012 & 0.608 & 0.590 & 0.415 & 0.322 & 0.270 & 0.022 & 0.866 & 0.863 & 0.822 & 0.381 & 0.362 & 0.228 & 0.464 & 0.470 & 0.406 \\
InternVL3 & \best{0.657} & \best{0.656} & \best{0.315} & 0.633 & 0.637 & 0.450 & \best{0.393} & \best{0.384} & \best{0.090} & 0.770 & 0.763 & 0.693 & 0.385 & 0.381 & 0.231 & 0.307 & 0.332 & 0.227 \\
\bottomrule
\end{tabular}%
} % End resizebox
\vspace{2mm} % Adjust vertical space after table if needed
\caption{Performance Comparison of Models Across Datasets and Prompting Regimes (Sorted by Class Count). Best score within each regime (Supervised, Zero Shot, One Shot) is \textbf{bolded}. Overall best score is \textbf{\underline{underlined}}.}
\label{tab:model_comparison_final_highlighted} % Changed label again
\end{table*}

We evaluated the performance of a baseline CNN and several state-of-the-art VLMs under zero-shot and one-shot prompting regimes across six distinct histopathology classification tasks. The detailed performance metrics (Accuracy, Macro F1-score, Cohen's Kappa) are presented in Table~\ref{tab:model_comparison_final_highlighted}.

Overall, the supervised CNN model consistently achieved the highest performance across most datasets, establishing a strong benchmark, particularly on Bladder Grade ($\kappa$=0.985), Bladder Tumor ($\kappa$=0.920), Matek \etal ($\kappa$=0.748), Meng \etal ($\kappa$=0.830), and PanNuke ($\kappa$=0.440). VLMs in the zero-shot setting generally struggled, often performing near or below chance levels, indicated by Kappa values close to or below zero (e.g., GPT 4.1 on Bladder Tumor: $\kappa$=0.005; Gemini 2.5 Pro on Bladder Grade: $\kappa$=-0.004).

\textbf{One-shot prompting significantly boosts VLM performance.} Comparing the zero-shot and one-shot results for the four VLMs across all six datasets reveals a substantial improvement when a single example per class is provided, particularly evident in the Cohen's Kappa scores. For instance, GPT 4.1's Kappa on Bladder Grade improves from 0.040 (zero-shot) to 0.393 (one-shot), and InternVL3's Kappa on Bladder Tumor increases from 0.235 to 0.315. To statistically validate this observation, we performed a \emph{one-sided paired Wilcoxon signed-rank test} comparing the zero-shot and one-shot Kappa scores across all VLM-dataset pairs (N=24). The results confirm a highly significant improvement with one-shot prompting ($p \approx 1.005 \times 10^{-5}$), demonstrating the effectiveness of minimal in-context learning for enhancing VLM performance on these pathology tasks.

\textbf{LLaVA-Med \cite{li2023llavamed} struggles with task comprehension.} Although LLaVA-Med was included in our initial evaluation due to its medical domain adaptation, its results are omitted from Table~\ref{tab:model_comparison_final_highlighted}. Under both zero-shot and one-shot prompting, it consistently failed to perform the classification task effectively, often failing to adhere to the class choices provided, providing blank responses in some cases and for the datasets that it was able to provide relevant output it performed poorly, for instance on Bladder Grade its Cohen's Kappa score was -0.005, similar trend was followed for other datasets. This suggests its general medical pre-training does not directly equip it for specific cellular/tissue morphology classification via simple prompting and highlights the need for more specialized adaptation or fine-tuning for such tasks.

\textbf{Performance does not strictly correlate with VLM scale.} Interestingly, larger models did not universally outperform smaller ones in our one-shot and zero-shot evaluations. For instance, on the Bladder Grade task, the smaller GPT 4.1 Mini ($\kappa$=0.515) and InternVL3 ($\kappa$=0.450) achieved higher agreement than both the larger GPT 4.1 ($\kappa$=0.393) and Gemini 2.5 Pro ($\kappa$=0.415). Similar trends were observed on Bladder Tumor (InternVL3 $\kappa$=0.315 outperformed others) and PanNuke (GPT 4.1 Mini $\kappa$=0.374 was highest). While factors like specific model architectures and pre-training data mixtures play a role, these results suggest that simply scaling model size may not guarantee optimal performance on specialized tasks like histopathology classification, highlighting the potential importance of architectural choices or training data composition even within large foundation models.

\textbf{VLMs generally achieve better results on datasets with larger image dimensions.} Comparing performance across datasets with varying average image sizes (Table~\ref{tab:datasets}) reveals a tendency for VLMs to perform better on images with larger pixel dimensions in the one-shot setting. For example, the highest Kappa scores were achieved on BCCD (Avg. 320x240 pixels; best VLM $\kappa$=0.893) and Bladder Grade (Avg. 280x280 pixels; best VLM $\kappa$=0.515). In contrast, performance was generally lower on datasets with smaller average sizes like PanNuke (Avg. 46x46 pixels; best VLM $\kappa$=0.374) and Bladder Tumor (Avg. 64x64 pixels; best VLM $\kappa$=0.315). This aligns with the observation that current general VLM visual encoders might be less adept at capturing the fine-grained details necessary for classification in lower-resolution images compared to tasks involving larger, more distinct objects or scenes.

\textbf{VLMs show potential, outperforming CNN on a specific task.} While the supervised CNN was dominant overall, VLMs demonstrated significant potential on the BCCD dataset. The one-shot GPT 4.1 model achieved the top performance (Acc: 0.920, F1: 0.920, $\kappa$=\textbf{0.893}), surpassing the baseline CNN ($\kappa$=0.615). Several other VLMs also outperformed the CNN on this task in the one-shot setting (e.g., Gemini 2.5 Pro: $\kappa$=0.822). While potentially influenced by the CNN's small training data size for this specific dataset, this result underscores the potential of VLMs replacing CNNs in medical field. It suggests that by leveraging vast prior knowledge, VLMs can achieve high accuracy with minimal examples (one-shot) and also have good performance in zero shot settings, hinting at their promise as potent tools for rapid hypothesis testing or as adaptable 'digital pathology assistants' in settings where extensive dataset curation for supervised training is impractical.
\section{Conclusion}
This study benchmarked the performance of contemporary Vision-Language Models (VLMs), including GPT-4.1, Gemini 2.5 Pro, and InternVL, on diverse histopathology classification tasks using zero-shot and one-shot in-context learning (ICL). Our findings demonstrate that while one-shot prompting significantly enhances VLM performance over zero-shot ($p \approx 1.005 \times 10^{-5}$ based on Kappa scores), indicating the potential of ICL for rapid adaptation, current general-purpose VLMs generally lag behind traditional supervised CNNs trained on these specific tasks. Performance did not strictly correlate with model scale, and VLMs tended to perform better on datasets with larger image dimensions. Notably, VLMs outperformed baseline CNN results on the BCCD datasets highlighting the potential of VLMs in pathological tasks. Overall, this study highlights both the promise of ICL for rapidly applying VLMs to pathology and the current performance gap compared to specialized supervised models. Future work should focus on developing VLMs with enhanced visual understanding tailored to cellular and tissue morphology, potentially through domain-specific pre-training or architectural adaptations, to bridge this gap for reliable diagnostic support.

{
    \small
    \bibliographystyle{ieeenat_fullname}
    \bibliography{main}
}


\end{document}